\pdfoutput=1
\documentclass[letterpaper]{article}
\usepackage[preprint]{aaai2027}
\usepackage[hyphens]{url}
\usepackage{graphicx}
\usepackage{natbib}
\usepackage{caption}
\usepackage{algorithm}
\usepackage{algorithmic}
\usepackage{booktabs}
\usepackage{amsmath}
\usepackage{amssymb}
\usepackage{multirow}


\newcommand{\figref}[1]{Fig.~\ref{#1}}
\newcommand{\tabref}[1]{Table~\ref{#1}}

\newcommand{\method}{ArtChart}
\newcommand{\bench}{ArtChart-Bench}
\newcommand{\eval}{ArtChart-Eval}

\title{\method: Faithful Artistic Chart Generation with Integrated Text Rendering}

\author{
 Meijia Huang \quad
 Yingjie Yin \quad
 Shihao Wang \quad
 Chenguang Ma
\vspace{1mm}
\\
\fontsize{11pt}{13pt}\selectfont{Ant Group}\\
{\tt\small
\{huangmeijia.hmj, gaoshi.yyj, shihao.wsh, chenguang.mcg\} \\
\tt\small@antgroup.com}
}
\affiliations{}

\begin{document}

\maketitle

\begin{figure*}[t]
\centering

\includegraphics[width=0.96\textwidth]{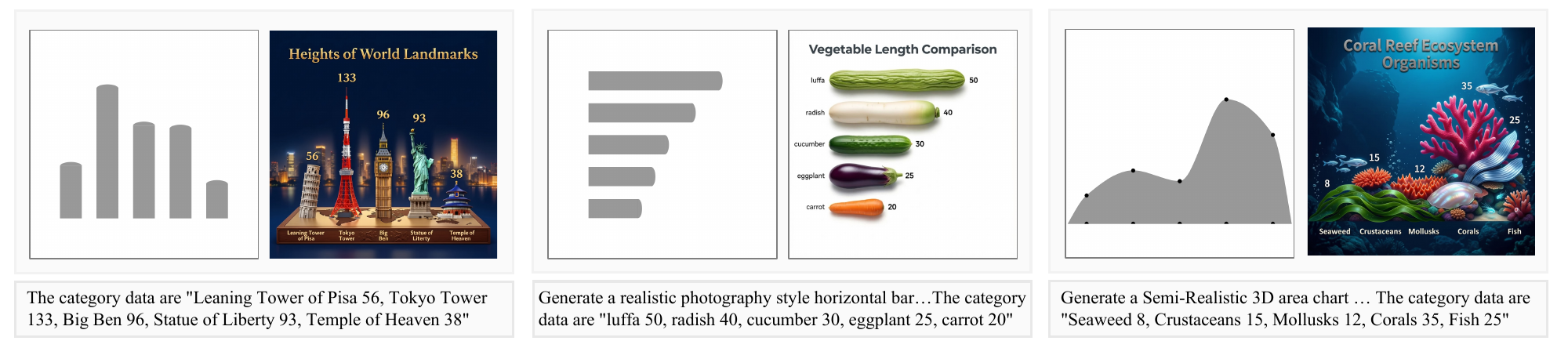}
\caption{Examples of artistic charts generated by \method. }
\label{fig:teaser}
\vspace{-10pt}
\end{figure*}

\begin{abstract}
Artistic charts combine data visualization with expressive marks, textures, and typography, but they are difficult for image generators: an output is useful only when its stylization preserves chart geometry, exact in-image text, and the semantic binding between labels and marks. We introduce \method{}, a framework for faithful artistic chart generation with integrated text rendering. Given a structured chart specification and an artistic prompt, \method{} first renders a text-free grayscale layout that encodes the target chart geometry, then trains a chart-specific control module to preserve mathematical structure. To address the remaining text and layout errors, we further refine the generation policy through GRPO-based reinforcement learning with OCR-based text rewards, VLM-based layout rewards, and aesthetic rewards. A multi-expert distillation stage reconciles these objectives by distilling single-reward experts into one balanced generation policy. We also construct \bench{}, a bilingual 2K-prompt benchmark covering four chart types, controlled value distributions, diverse label/value formats, and 15 artistic styles, together with \eval{}, a six-axis evaluation protocol measuring mathematical logic, text accuracy, text layout, aesthetics, instruction following, and readability. Experiments on \bench{} show that \method{} consistently outperforms prompt-only, image-editing, and generic ControlNet baselines, with the largest gains on mathematical fidelity and label-layout binding while maintaining competitive visual quality. These results suggest that artistic chart generation should be evaluated as reliable visual communication rather than as generic stylized image synthesis.
\end{abstract}

\section{Introduction}

Statistical charts are precise instruments for communicating data. Tools such as Matplotlib, ECharts, and Vega-Lite produce reproducible geometry, but their visual vocabulary is usually limited to conventional marks and layouts. Artistic charts pursue a complementary goal: they use expressive marks, textures, typography, and thematic backgrounds to make data more memorable in journalism, education, advertising, and presentations. Such charts are expensive because designers must preserve data fidelity while manually shaping marks, placing labels, and repairing readability issues.

\begin{figure}[t]
\centering

\includegraphics[width=0.85\columnwidth]{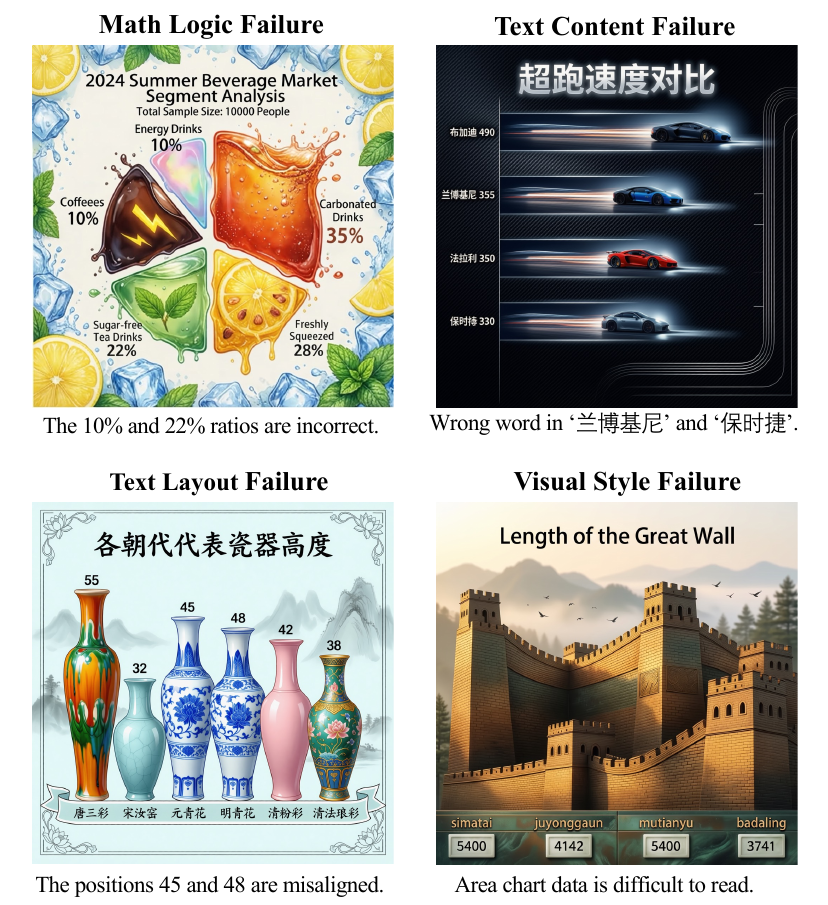}
\caption{Failure taxonomy.}
\label{fig:failures}
\vspace{-10pt}
\end{figure}

Modern text-to-image (T2I) models have made rapid progress in style synthesis, instruction following, and visual text rendering \citep{qwenimage2025,ernieimage2026}. This raises the possibility of automatically generating artistic charts. However, chart generation is not ordinary image synthesis. A visually pleasing output is invalid if a bar has the wrong height, a pie sector has the wrong angle, a digit is hallucinated, or a correct value is placed above the wrong category. Through pilot experiments, we summarize four recurring failures: mathematical-logic distortion, text-content errors, text-layout mismatch, and visual-style/readability failure (Fig.~\ref{fig:failures}).

We study \emph{artistic chart generation with integrated text rendering}. Given a chart specification and a style prompt, a model should generate a raster image whose background is stylized, whose geometric marks are artistically transformed (e.g., bars as stacked books or scallions), and whose title, category labels, and value labels are rendered in the correct semantic locations. This differs from pictorial chart generation, which focuses on replacing marks with semantic objects but does not systematically render chart text \citep{xiao2026chartist,sun2026ssalign}, and from creative table visualization, which generates infographic-like layouts through multi-stage MLLM workflows \citep{liu2026showtable}.

Our method, ArtChart, adopts a three-stage training pipeline to jointly address structural fidelity, text accuracy, and visual aesthetics. First, we render the input data into a text-free grayscale chart and train a chart-specific ControlNet, enabling explicit structural grounding for bar, horizontal-bar, pie, and area charts. Second, since the remaining dominant errors are text and layout failures, we freeze the mathematical control module and train DiT LoRA adapters using reinforcement learning with OCR text accuracy and VLM text-position rewards, progressively refining text correctness and spatial layout alignment. Third, we incorporate an aesthetic reward and address inter-reward conflicts through a multi-expert distillation strategy: three single-reward experts are trained independently to their respective performance ceilings, then distilled into a unified student model using dense teacher velocity fields. This design keeps mathematical control explicit while improving text rendering and visual aesthetics through on-policy alignment.

To support systematic evaluation, we introduce \bench, a bilingual benchmark with 1K English and 1K Chinese prompts, spanning four chart types, diverse label formats, and 15 artistic styles, with structured data specifications and diagnostic metadata for fine-grained analysis. We further design \eval, a six-axis evaluation suite covering mathematical logic, text accuracy, text layout, aesthetics, instruction following, and readability, enabling unified and systematic comparison across T2I, image-editing, and controllable generation models under a single prompt-output protocol.

Our contributions are:
\begin{itemize}
    \item We define artistic chart generation as a faithful image-generation task requiring numerical geometry, integrated text rendering, text-layout binding, and artistic stylization.
    \item We propose ArtChart, combining a plug-and-play math control module, RL learning with text, layout, and aesthetic rewards, and a multi-expert distillation strategy to resolve inter-reward conflicts for balanced optimization.
    \item We construct \bench, a 2K bilingual benchmark with controlled data, text, style, chart-type distributions, diagnostic metadata, and a failure taxonomy tailored to artistic charts.
    \item We provide \eval, a six-axis automatic evaluation protocol and a controlled experimental design over T2I, editing, controllable-generation, and ablation baselines.

\end{itemize}

\section{Related Work}

\begin{figure*}[t]
\centering

\includegraphics[width=0.80\textwidth]{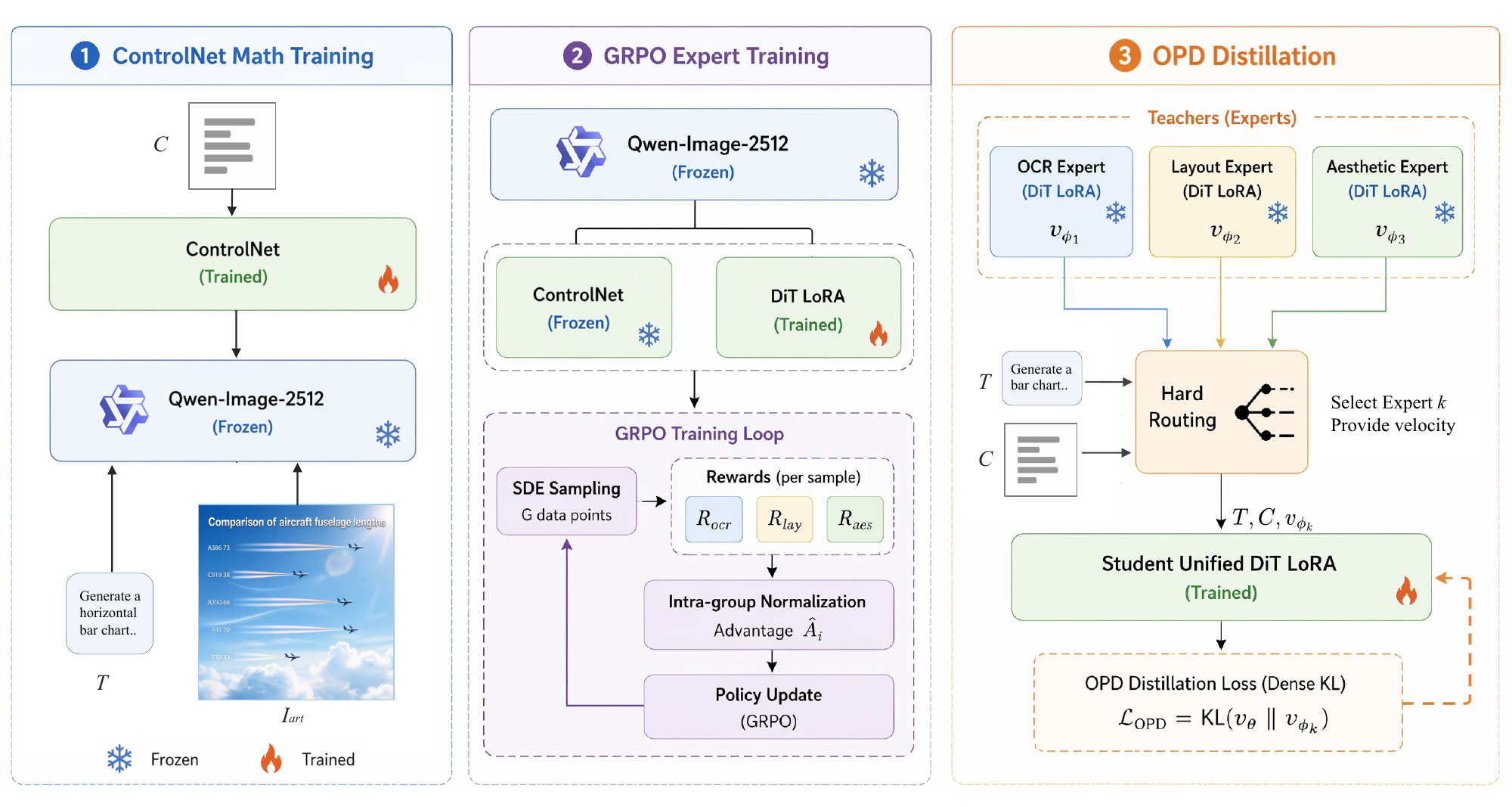}
\caption{\method{} framework. The figure illustrates the three-stage training pipeline of our method: (1) a mathematical control module trained to ensure chart structural and logical fidelity; (2) a reinforcement learning stage optimizing text accuracy, text layout, and aesthetics through complementary reward signals; and (3) a multi-expert distillation stage that jointly improves all three aspects through specialized expert optimization.}
\label{fig:framework}
\vspace{-10pt}
\end{figure*}

\subsection{Chart and Infographic Generation}

Classical visualization systems such as Matplotlib, D3.js, and ECharts provide deterministic data-to-geometry mappings and editable outputs \citep{hunter2007matplotlib,bostock2011d3,li2018echarts}. Recent natural-language visualization tools such as LIDA and Chat2VIS further translate user intent into visualization code \citep{dibia2023lida,maddigan2023chat2vis}. These systems are strong for conventional charts because correctness is backed by executable specifications, but they are less suited to fully rasterized artistic charts where marks, textures, typography, and backgrounds are synthesized jointly.

A related line of work studies pictorial and stylized charts. ChartSpark embeds semantic context into chart foreground or background and evaluates data distortion \citep{xiao2024chartspark}; viz2viz decomposes and recomposes chart marks through diffusion pipelines \citep{wu2023viz2viz}; ChartEditor supports human-AI paired pictorial chart authoring \citep{yan2025charteditor}. Recent ChArtist and SSAlign improve structure-subject alignment for pictorial chart generation \citep{xiao2026chartist,sun2026ssalign}. However, these methods mainly focus on replacing or transforming chart marks while preserving their spatial extent, and they do not address integrated chart-text generation. In contrast, \method{} requires titles, category labels, and value labels to be rendered inside the generated image and treats label-value binding as a core metric.

ShowTable converts markdown tables into creative table visualizations through multi-stage rewriting, generation, reflection, and refinement \citep{liu2026showtable}. However, it does not address artistic transformation of chart primitives that encode numerical values. IGenBench evaluates text-to-infographic reliability through atomic verification questions \citep{tang2026igenbench}, while ChartQA and MatplotBench evaluate chart understanding or chart-code generation \citep{masry2022chartqa,yang2024matplotbench}. \bench{} differs by targeting generated artistic chart images in which quantitative geometry, visual text, and artistic mark deformation are produced by image models and can fail independently.

\subsection{Visual Text Rendering}

Accurate visual text rendering is essential for chart generation because a single wrong digit, unit, or category name can invalidate the image. GlyphByT5, AnyText, TextDiffuser, and EasyText improve text rendering with character-aware encoders, glyph priors, or layout control \citep{liu2024glyphbyt5,tuo2024anytext,chen2023textdiffuser2,liu2025easytext}. TextPecker and TextAlign further introduce reward-based or hierarchical alignment signals for visual text generation \citep{zhu2026textpecker,cui2026textalign}. Artistic chart text imposes an additional constraint beyond ordinary poster or scene text: each string must be both correct and semantically attached to the corresponding mark. We therefore evaluate text accuracy and text layout as separate dimensions.

\subsection{Controllable Generation and RL Alignment}
ControlNet conditions image generation on spatial layouts such as edges, depth, and masks \citep{zhang2023controlnet}, with later variants improving control consistency \citep{li2024controlnetpp}. For artistic charts, the condition is a text-free grayscale chart whose mark geometry encodes numerical values, motivating our chart-specific control module and generic-ControlNet baselines in \bench{}.

Reinforcement learning has recently been used to align diffusion and flow-based generators with non-differentiable rewards. DDPO and DPOK optimize diffusion models with online reward feedback \citep{black2024ddpo,fan2024dpok}, while Flow-GRPO adapts group-relative policy optimization to flow-matching models through stochastic sampling paths \citep{liu2025flowgrpo}. Multi-objective alignment remains challenging because rewards for text accuracy, layout, aesthetics, and readability can interfere. Flow-OPD addresses such conflicts by training single-reward experts and distilling them into one student policy \citep{fang2026flowopd}, and POCA studies Pareto-aware sample selection for visual text alignment \citep{fan2026poca}. We build on these ideas in a structured chart-generation setting where the competing objectives can be diagnosed separately.

\section{Task, Taxonomy, and Benchmark}

\textbf{Task definition.}
An input instance is $(T,D,P)$, where $T\in\{\mathrm{bar},\mathrm{hbar},\mathrm{pie},\mathrm{area}\}$ is a chart type, $D=\{(c_i,v_i)\}_{i=1}^{n}$ is a one-dimensional categorical dataset, and $P$ is an artistic prompt describing the style, theme, mark metaphor, title, and layout. The output image $I$ should satisfy four requirements: (1) marks encode $D$ according to $T$; (2) marks and background are artistically stylized and coherent with $P$; (3) the title, categories, and values are rendered correctly; and (4) each category/value is placed at the correct semantic location.

\textbf{Failure taxonomy.}
We group failures into four classes. \emph{ML} (math logic): proportion distortion, order errors, and wrong mark counts. \emph{TC} (text content): missing, duplicated, wrong, or garbled characters. \emph{TL} (text layout): label-value mismatch, overlap, or position drift. \emph{VS} (visual style): prompt-style mismatch or over-stylization that makes marks unreadable.

\textbf{Benchmark construction.}
\bench{} contains 2K evaluation prompts: 1K English and 1K Chinese, each with 250 samples per chart type. Category counts follow a long-tailed distribution from 2 to 10: 2 (3\%), 3 (15\%), 4 (17\%), 5 (20\%), 6 (18\%), 7 (12\%), 8 (8\%), 9 (4\%), and 10 (3\%). Value orders include unordered (60\%), increasing (15\%), decreasing (15\%), and all-equal (10\%) sequences; value magnitudes include large-difference (40\%), small-difference (40\%), extreme-value (10\%), and zero/tiny-value (10\%) cases. Category text covers pure words, digit-word mixtures, alphanumeric labels, pure numbers, mixed-language strings, and symbols such as currency signs. Display values cover pure numbers (33\%), number+unit (33\%), and number+symbol (34\%). We use 15 style families from minimalist to highly complex.

\textbf{Prompt and specification schema.}
Each benchmark instance stores both natural-language and structured fields. The natural-language prompt is a compact paragraph specifying chart type, topic, artistic theme, mark metaphor, color palette, title, and data list. The structured record stores \texttt{id}, \texttt{language}, \texttt{chart\_type}, \texttt{title}, \texttt{data}, \texttt{raw\_values}, \texttt{display\_values}, \texttt{style}, and \texttt{diagnostic\_tags}. This dual representation is important: generators receive prompts in the form they naturally support, while evaluators use structured ground truth to compute data and text scores. The released benchmark will include the grayscale condition $G$ for methods that support image conditions, while prompt-only, editing, and controllable-generation methods are all evaluated from the same final image. Diagnostic tags support fine-grained analysis by language, chart type, value pattern, label format, style family, and difficulty. We will release the benchmark and evaluation code for research use under a permissive license.

\begin{table}[t]
\centering
\small
\setlength{\tabcolsep}{3pt}
\begin{tabular}{lll}
\toprule
Field & Purpose & Example diagnostic use \\
\midrule
Language & English/Chinese split & multilingual text gap \\
Chart type & bar/hbar/pie/area & geometry-specific failures \\
Value pattern & order/magnitude tag & ratio and order stress tests \\
Text format & label/value format & OCR and glyph failures \\
Style family & 15 visual styles & readability vs. aesthetics \\
Difficulty & hard-case tag & long labels, tiny values \\
\bottomrule
\end{tabular}
\caption{Core metadata in \bench.}
\label{tab:metadata}
\vspace{-10pt}
\end{table}

\begin{figure}[t]
\centering

\includegraphics[width=0.98\columnwidth]{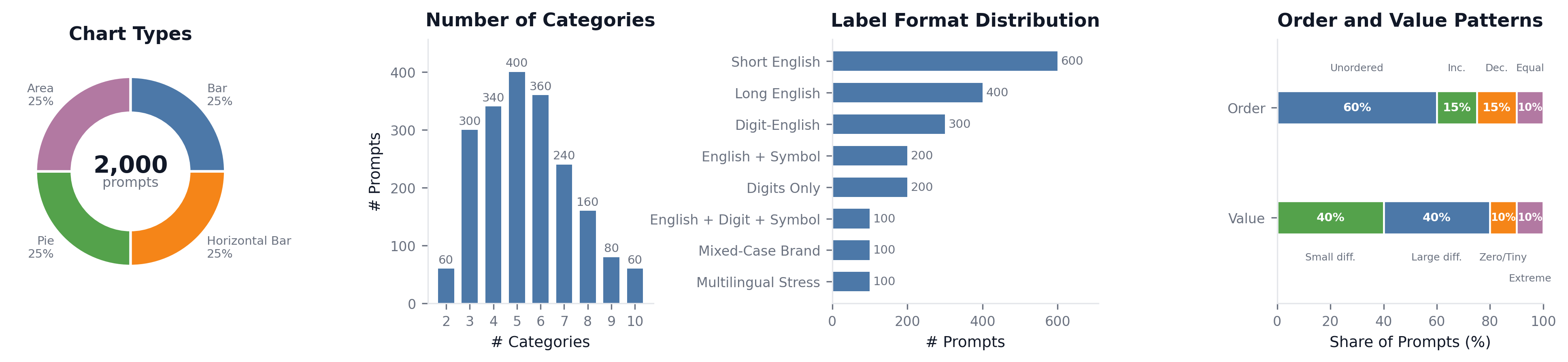}
\caption{\bench{} data distribution.}
\label{fig:bench}
\vspace{-10pt}
\end{figure}

\begin{figure}[t]
\centering

\includegraphics[width=0.98\columnwidth]{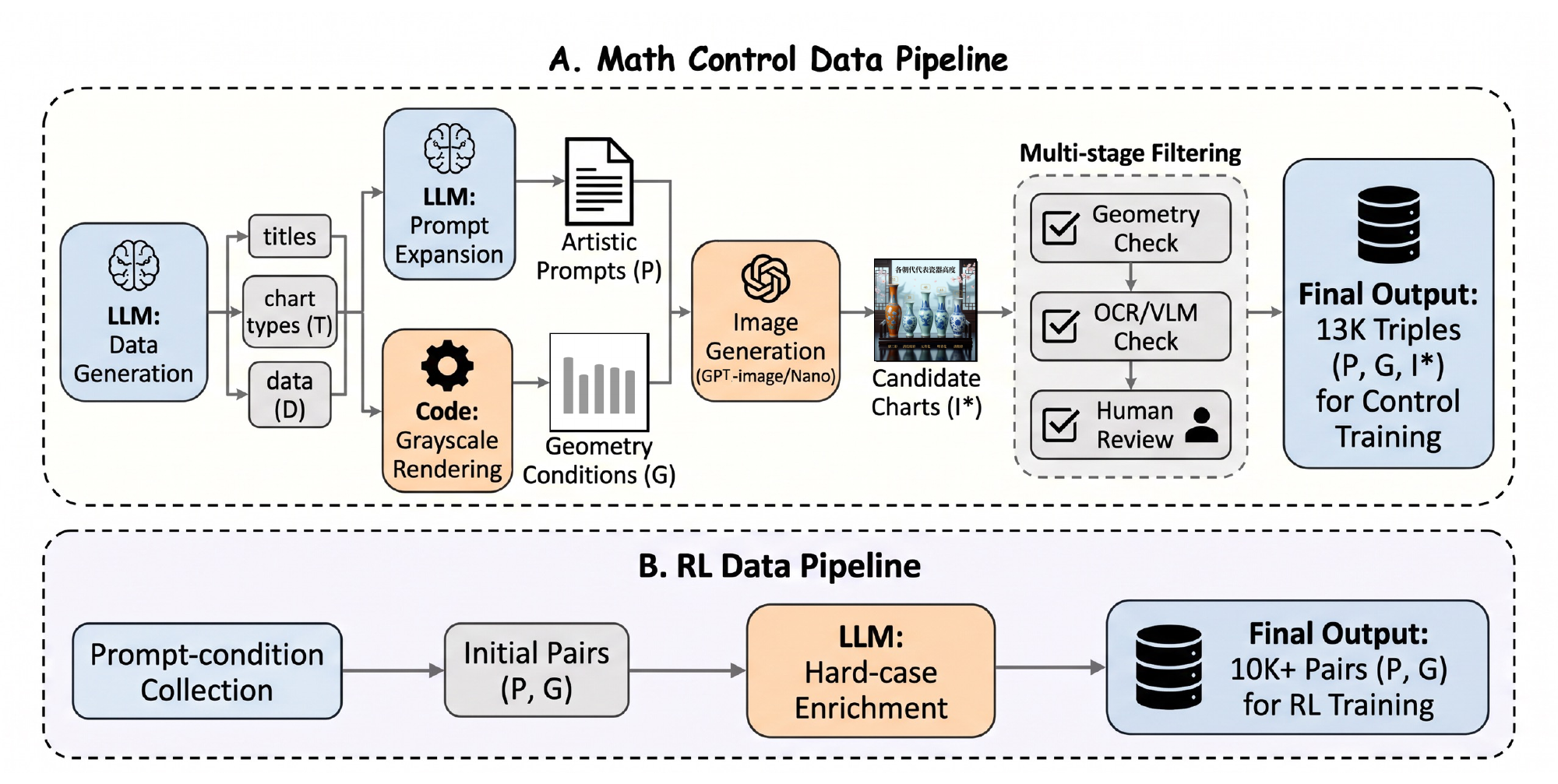}
\caption{Training Data Construction Process.}
\label{fig:train_data}
\vspace{-8pt}
\end{figure}

\textbf{Training data.}
Training the math control module uses about 13K triples $(P,G,I^\star)$ shown in Fig.~\ref{fig:train_data}. We first use an LLM to construct chart titles and data $D$ covering diverse topics and numerical patterns. We then use a chart-type-aware prompt-expansion template to convert each title, chart type, and data specification into a complete generation prompt $P$, including descriptions of style, background, color palette, artistic mark metaphor, and text appearance. Given the chart type $T$ and data $D$, we render a deterministic text-free grayscale chart $G$ that contains only mark geometry, with no axes, ticks, legends, or text. The target image $I^\star$ is generated by GPT-Image and Nano teacher generators conditioned on $(P,G)$. Since these models still produce many invalid artistic charts, we clean the generated images with a multi-stage filtering pipeline: geometry filtering against $G$ to remove data-distortion errors, OCR/VLM filtering for text errors and layout confusion, and human review for unreadability or poor visual quality. The RL stage uses 10K+ prompt-condition
pairs $(P,G)$, enriched with hard cases such as long labels and categories/values that are easy to swap. In later construction rounds, we enforce a stable layout convention, with categories near the base and values near mark endpoints, because pilot SFT experiments showed that inconsistent teacher text placement led to unstable generated layouts.

\textbf{Quality control.}
The benchmark and training data follow separate quality-control procedures. Evaluation prompts are manually checked for unambiguous data, valid style descriptions, and deterministic chart type. Training images are more aggressively filtered because noisy teacher images can teach wrong geometry or unstable layout. We first compare artistic mark geometry with the grayscale condition; then run OCR and VLM layout checks; finally, human reviewers discard examples with text hallucination, label-value swaps, implausible mark proportions, or over-stylized marks that cannot be decoded as charts. This separation avoids leaking model-specific teacher artifacts into the benchmark while still allowing large-scale training data construction.

\section{Method}

\subsection{Chart-Specific Mathematical Control Module.}

We use Qwen-Image-2512 as the T2I backbone because it provides strong multilingual text rendering and instruction following. Given $(T,D)$, we first render a grayscale condition $G$ with deterministic Python chart code. For vertical and horizontal bars, $G$ encodes bar height or length; for pie charts, sector angles; for area charts, filled regions. The condition contains no title, category labels, values, axes, or ticks. We train a chart-specific ControlNet branch on $(P,G,I^\star)$ while freezing the base generator:
\begin{equation}
\mathcal{L}_{\mathrm{ctrl}} =
\mathbb{E}_{t,\epsilon,I^\star}
\left\|v_{\theta,\phi}(x_t,t,P,G)-v^\star_t\right\|_2^2 .
\end{equation}
Here $x_t$ is the noisy latent at timestep $t$, $v^\star_t$ is the target flow velocity, $\theta$ denotes the frozen base model, and $\phi$ denotes the trainable ControlNet parameters. The grayscale condition deliberately contains no text, so the controller learns mathematical structure rather than copying chart scaffolding. In practice, this stage uses about 13K filtered $(P,G,I^\star)$ triples. The strongest improvement is on MathLogic, but we also observe that consistent teacher layouts help the ControlNet stage learn coarse text-placement conventions.

\subsection{GRPO for Text and Aesthetic Alignment.}

After ControlNet training, the dominant failures are no longer global mark placement but text corruption and label-value mismatch. We therefore freeze both the ControlNet and the
base model, attach a LoRA adapter to the base DiT, and optimize only this DiT LoRA. The key implementation point is that RL samples must still be generated with the trained ControlNet attached; otherwise the policy would optimize text on a distribution different from final inference.

Following Flow-GRPO, we convert the deterministic flow ODE into an equivalent SDE with the same marginal distribution:
\begin{equation}
\resizebox{0.95\linewidth}{!}{$
dx_t = \left[v_\theta(x_t,t,c,G) + \frac{\sigma_t^2}{2t}\big(x_t-(1-t)v_\theta(x_t,t,c,G)\big)\right]dt + \sigma_t dW_t,
$}
\label{eq:sde}
\end{equation}
where $\sigma_t=a\sqrt{t/(1-t)}$ and $a=0.7$. This stochastic path gives a Gaussian transition density and enables policy-gradient optimization. For each prompt-condition pair, we sample a group of $M$ candidates $\{x_0^{(i)}\}_{i=1}^{M}$, compute rewards $R_i$, and normalize advantages within the group:
\begin{equation}
A_i=\frac{R_i-\mathrm{mean}(\{R_j\}_{j=1}^{M})}{\mathrm{std}(\{R_j\}_{j=1}^{M})}.
\end{equation}
The LoRA policy is updated with a clipped GRPO objective:
\begin{equation}
\resizebox{0.95\linewidth}{!}{$
\mathcal{L}_{\mathrm{GRPO}} =
-\mathbb{E}\left[
\min\left(\rho_i A_i,\mathrm{clip}(\rho_i,1-\epsilon,1+\epsilon)A_i\right)
-\beta D_{\mathrm{KL}}(\pi_\theta\|\pi_{\mathrm{ref}})
\right],
$}
\label{eq:grpo}
\end{equation}
where $\rho_i$ is the likelihood ratio between the current and old policies computed from the SDE transitions. We use denoising reduction: 10 denoising steps during RL training, 40 steps for full inference, and 4 steps with Lightning LoRA for accelerated inference.

We use two text rewards. The OCR reward extracts text from $I$ and matches it to the target title, categories, values, and units with Hungarian matching under normalized edit distance:
\begin{equation}
\resizebox{0.95\linewidth}{!}{$
R_{\mathrm{ocr}}(x_0,c)=\mathrm{clip}\left(
\frac{\sum_{(t,p)\in\mathcal{M}}\mathrm{NED}(t,p)}
{\max(|T|,|P|)}-\mathrm{Penalty},0,1\right).
$}
\label{eq:rocr}
\end{equation}
Here $T$ is the target text set, $P$ is the OCR-detected text set, and $\mathcal{M}$ is the Hungarian matching. The layout reward checks semantic attachment. Each data point contributes two slots, category and value. A VLM judges whether each slot has correct content and correct position/order, giving
\begin{equation}
R_{\mathrm{lay}}(x_0,c)=\frac{\#\mathrm{correct\ slots}}{2n}.
\label{eq:rpos}
\end{equation}
For bar and area charts, positions are checked relative to the ordered marks; for pie charts, positions are checked relative to sectors. This reward captures errors that OCR alone cannot detect, such as correct strings attached to the wrong mark.

\subsection{Multi-Expert Distillation.}

We train a third Flow-GRPO expert with PickScore reward $R_{\mathrm{aes}}=\mathrm{PickScore}(I,P)$ \citep{kirstain2023pickscore}. Directly optimizing a weighted sum $w_1R_{\mathrm{ocr}}+w_2R_{\mathrm{lay}}+w_3R_{\mathrm{aes}}$ creates a seesaw effect: improving aesthetics may reduce text accuracy, while aggressively optimizing OCR often produces plain or rigid layouts. We therefore use OPD.

\emph{Stage 1: single-reward experts.}
We train three LoRA experts $\phi_{\mathrm{ocr}}$, $\phi_{\mathrm{lay}}$, and $\phi_{\mathrm{aes}}$ with Flow-GRPO under $R_{\mathrm{ocr}}$, $R_{\mathrm{lay}}$, and $R_{\mathrm{aes}}$, respectively. Each expert is optimized until its own validation reward saturates.

\emph{Stage 2: on-policy distillation.}
A student LoRA samples trajectories on policy. A hard router selects one teacher expert for each sample, and the selected expert provides a reference velocity field. Because student and teacher transitions share covariance, the reverse KL reduces to a dense velocity loss:
\begin{equation}
\resizebox{0.9\linewidth}{!}{$
D_{\mathrm{KL}}(\pi_\theta\|\pi_{\phi_k}) =
\frac{\Delta t}{2}\left(\frac{\sigma_t(1-t)}{2t}+\frac{1}{\sigma_t}\right)^2
\|v_\theta-v_{\phi_k}\|_2^2 .
$}
\label{eq:opdkl}
\end{equation}

The final OPD LoRA is used as the default \method{} model in the benchmark.

\section{Evaluation Protocol}

\begin{figure*}[t]
\centering

\includegraphics[width=0.96\textwidth]{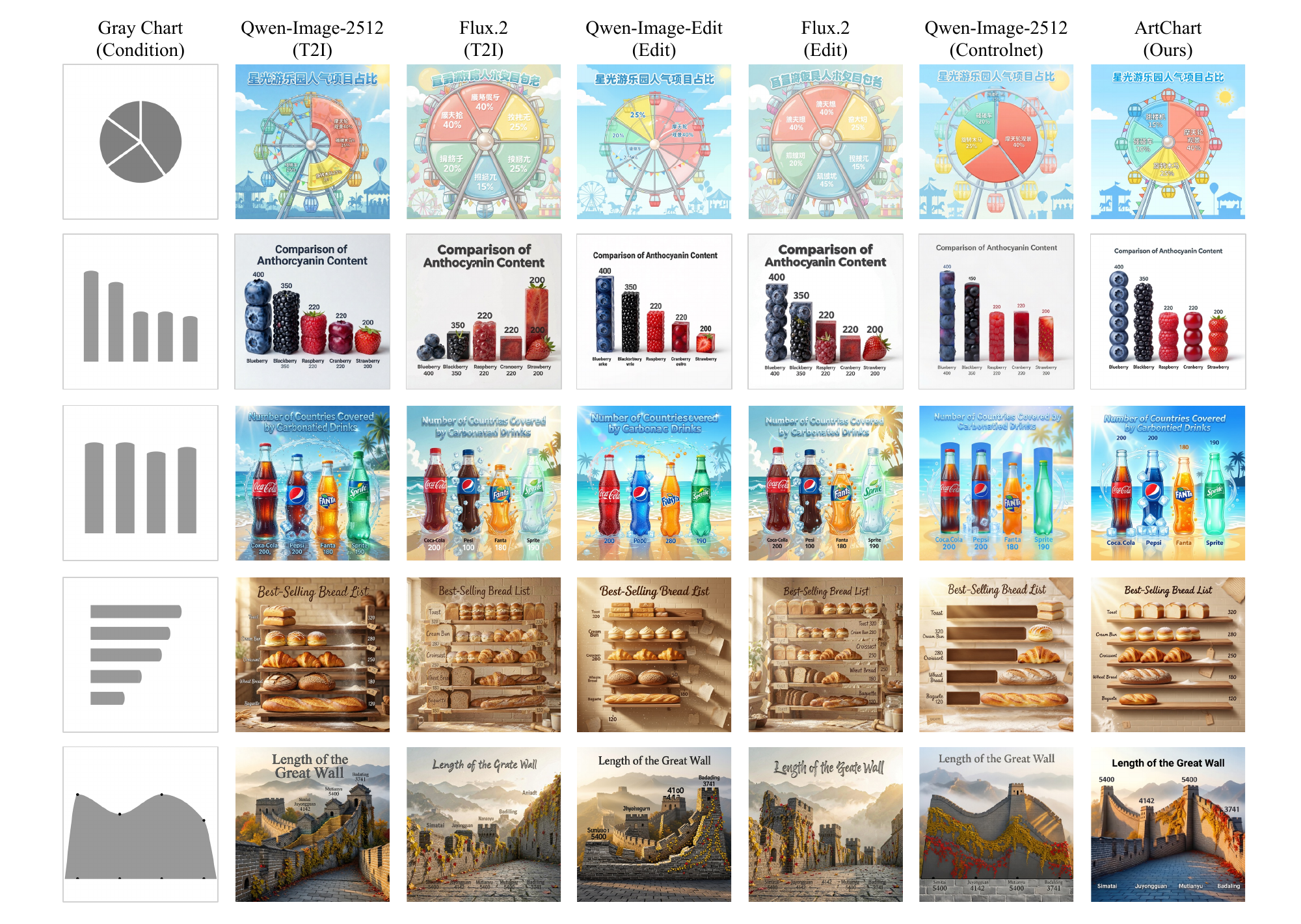}
\caption{Qualitative comparison. Compared with prompt-only, editing, and generic open ControlNet baselines, \method{} better preserves chart geometry, renders readable text, and maintains label-value alignment under artistic styles.}
\label{fig:qualitative}

\end{figure*}

\eval{} reports six scores in $[0,10]$. The evaluator uses the prompt/specification and generated image; methods that consume grayscale conditions do so only at generation time, not as a separate leaderboard.

\begin{table}[t]
\centering
\small
\setlength{\tabcolsep}{3pt}
\begin{tabular}{lll}
\toprule
Metric & Main signal & Typical failures \\
\midrule
MathLogic & mark ratios/order/count & wrong height, angle, area \\
TextAcc & OCR + edit distance & missing digits, wrong units \\
LayoutPos & VLM slot checking & label-value swap, drift \\
Aesthetic & preference model & weak style, poor composition \\
InstrFollow & atomic VLM checks & missing theme/metaphor \\
Readability & VLM usability check & over/under stylization \\
\bottomrule
\end{tabular}
\caption{\eval{} dimensions.}
\label{tab:metrics}
\vspace{-10pt}
\end{table}

\textbf{Evaluation principles.}
\eval{} is built to serve a benchmark rather than a single model. We therefore avoid metrics that depend on hidden model internals or on a particular generation route. The same scoring code is applied to prompt-only T2I outputs, image-editing outputs, ControlNet outputs. We report all six component scores instead of only the average because artistic chart generation has meaningful trade-offs: a model may be visually attractive but mathematically wrong, or text-faithful but too plain to count as an artistic chart. The leaderboard therefore reports both a single aggregate and a score vector.

\textbf{MathLogic.}
We evaluate mathematical consistency with a text-masked VLM protocol. Instead of asking the Qwen3.5-397B-A17B VLM to directly output a score, we ask it to output a set of numerical ratios. OCR first detects all text boxes, which are covered by white masks to prevent the VLM from reading chart labels or values. The VLM then estimates the
visual mark proportions $\hat{\mathbf{r}}=\{\hat r_i\}_{i=1}^{m}$ in chart order. If $m\neq n$, the score is set to zero. Given normalized ground-truth ratios $\mathbf{r}=\{r_i\}_{i=1}^{n}$, we compute
\begin{equation}
\resizebox{0.85\linewidth}{!}{$
S_{\mathrm{ratio}}=
10\max\left(0,1-\frac{\max(0,\frac{1}{n}\sum_i|\hat r_i-r_i|-\tau)}
{\epsilon}\right),
$}
\end{equation}
where $\tau=0.05$ and $\epsilon=0.45$.

We further measure pairwise order consistency:
\begin{equation}
\resizebox{0.9\linewidth}{!}{$
S_{\mathrm{order}}=
10 \cdot
\frac{
\sum_{i<j}\mathbb{I}\left[|r_i-r_j|>\delta\right]
\mathbb{I}\left[(r_i-r_j)(\hat r_i-\hat r_j)>0\right]
}{
\sum_{i<j}\mathbb{I}\left[|r_i-r_j|>\delta\right]
}.
$}
\end{equation}
The final score is $S_{\mathrm{MathLogic}}
=0.8S_{\mathrm{ratio}}+0.2S_{\mathrm{order}}$.

\textbf{Text Accuracy.}
We evaluate chart text accuracy at the text-unit level, since OCR reading order is often unstable in artistic charts. Let $\mathcal{G}=\{g_i\}_{i=1}^{N}$ denote the ground-truth text units, including titles, category labels, and value labels, and let $\mathcal{P}=\{p_j\}_{j=1}^{M}$ denote the OCR-recognized text units. For each pair, we compute normalized edit similarity:
\begin{equation}
\resizebox{0.65\linewidth}{!}{$
\mathrm{sim}(g_i,p_j)=1-\frac{\mathrm{ED}(g_i,p_j)}{\max(|g_i|,|
p_j|)},
$}
\end{equation}
where $\mathrm{ED}(\cdot,\cdot)$ is the Levenshtein edit
distance.

We perform maximum-weight bipartite matching and keep matches above a threshold $\gamma$. The final score is a 10-point soft F1 score:
\begin{equation}
\resizebox{0.96\linewidth}{!}{$
S_{\mathrm{text}}=10\cdot\frac{2PR}{P+R},\quad
P=\frac{\sum_{(i,j)\in\mathcal{M}}\mathrm{sim}(g_i,p_j)}{M},\quad
R=\frac{\sum_{(i,j)\in\mathcal{M}}\mathrm{sim}(g_i,p_j)}{N}.
$}
\end{equation}
This order-free F1 formulation penalizes both missing and
hallucinated text while allowing minor OCR recognition errors.

\textbf{Text Layout.}
Layout evaluates the placement of chart text. A VLM checks each category and value label, and a text slot is counted as correct only if its content matches the reference and its position follows the expected chart order:
\begin{equation}
S_{\mathrm{layout}}=\frac{\#\mathrm{correct\ text\ slots}}
{2n}\times10.
\end{equation}

\begin{table}[t]
\centering
\small
\setlength{\tabcolsep}{2.3pt}
\begin{tabular}{lcccccc}
\toprule
Method & Math & Text & Layout & Aes. & IF & Read. \\
\midrule
Qwen-Image-2512 & 2.87 & 7.06 & 4.93 & 5.29 & 7.03 & 6.83 \\
FLUX.2-dev(T2I) & 1.23 & 5.88 & 4.70 & 5.17 & 6.36 & 5.96 \\
Qwen-Edit-2511 & 5.24 & 6.48 & 4.45 & 5.14 & 6.49 & 6.82 \\
FLUX.2-dev(Edit) & 3.07 & 6.00 & 4.88 & 5.22 & 6.63 & 6.02 \\
Qwen+CN-open & 7.48 & 8.05 & 5.17 & 5.08 & 7.83 & 7.27 \\
\textbf{\method} & \textbf{8.00} & \textbf{8.45} & \textbf{7.67} & \textbf{5.36} & \textbf{9.25} & \textbf{7.54} \\
\bottomrule
\end{tabular}
\caption{Main quantitative results on \bench{} (Chinese+English weighted average).}
\label{tab:main}
\vspace{-5pt}
\end{table}

\textbf{Instruction, readability, aesthetics.}
Instruction following checks chart type, theme, mark metaphor, title position, color palette, and decorative elements with atomic VLM questions. Readability penalizes both over-artistic outputs whose marks/text cannot be decoded and under-stylized outputs indistinguishable from plain charts. Aesthetic quality is scored by Aesthetic Predictor \citep{aestheticpredictor}, with VLM style consistency as auxiliary verification. Overall score is the unweighted average of the six axes.

\section{Experiments}

\textbf{Training configuration.}
ControlNet training uses the 13K filtered triples at $1024\times1024$ or $1024\times768$ resolution, with the base Qwen-Image-2512 weights frozen. Flow-GRPO uses 10K+ prompt-condition pairs, group size 16 for the main text/layout runs and 24 for OPD sampling, noise scale $a=0.7$, KL weight $\beta=5\times10^{-3}$, clipping threshold $\epsilon=10^{-4}$, and LoRA rank/scale $r=32,\alpha=32$. We freeze the chart ControlNet during all RL and OPD stages. This separation is deliberate: the ControlNet is responsible for mathematical geometry, while the DiT LoRA is responsible for text rendering, text placement, and visual style.

\textbf{Inference.}
The final model composes Qwen-Image-2512, the chart ControlNet, the OPD LoRA, and an optional Lightning LoRA. The method takes a prompt and a grayscale image as input. We use an LLM prompt expansion module to transform the structured data and content theme inputs into image-generation prompts, incorporating elements such as chart type, title, data list, visual metaphors, color schemes, and stylistic descriptions.

\textbf{Main Results.}
Table~\ref{tab:main} and Fig.~\ref{fig:qualitative} show that prompt-only T2I retains moderate aesthetics but fails on math, text, and layout. Editing improves structure but remains weak on text. Generic ControlNet improves math but is not tuned for artistic chart text. \method{} improves all six metrics.

\textbf{Per-chart-type analysis.}
Table~\ref{tab:pertype} shows that \method{} performs
best on bar and horizontal-bar charts. Area charts keep
strong math/text scores but lower readability, while pie
charts remain the hardest due to sector-angle and label-
placement challenges.

\begin{table}[t]
\centering
\small
\setlength{\tabcolsep}{2.5pt}
\begin{tabular}{lcccccc}
\toprule
Type & Math & Text & Layout & Aes. & IF & Read. \\
\midrule
Bar  & 8.79 & 8.47  & 7.89 & 5.41 & 9.58 & 9.27 \\
H-Bar & 8.67 & 8.46 & 7.86 & 5.19 & 9.60 & 8.21 \\
Area & 7.71 & 8.45 & 7.99 & 5.38 & 9.46 & 6.64 \\
Pie  & 6.82 & 8.42 & 6.94 & 5.45 & 8.36 & 6.05 \\
\bottomrule
\end{tabular}
\caption{Per-chart-type results of \method{}}
\label{tab:pertype}

\end{table}

\textbf{Component ablation.}
Table~\ref{tab:ablation} follows the module sequence. Chart-specific ControlNet raises mathematical logic and stabilizes layout because SFT learns spatial placement conventions. GRPO improves text and layout but slightly hurts aesthetics. OPD recovers aesthetics while retaining text/layout improvements.

\begin{table}[t]
\centering
\small
\setlength{\tabcolsep}{2.3pt}
\begin{tabular}{lcccccc}
\toprule
Configuration & Math & Text & Layout & Aes. & IF & Read. \\
\midrule
Qwen-Image-2512 & 2.87 & 7.06 & 4.93 & 5.29 & 7.03 & 6.83 \\
+ open CN & 7.48 & 8.05 & 5.17 & 5.08 & 7.83 & 7.27 \\
+ chart CN & \textbf{8.06} & 8.39 & 7.39 & 5.20 & 9.11 & 7.50 \\
+ GRPO text & 8.02 & \textbf{8.49} & 7.52 & 5.22 & 8.51 & \textbf{7.56} \\
+ OPD full & 8.00 & 8.45 & \textbf{7.67} & \textbf{5.36} & \textbf{9.25} & 7.54 \\
\bottomrule
\end{tabular}
\caption{Ablation of structural control, GRPO, and OPD.}
\label{tab:ablation}
\vspace{-4pt}
\end{table}

\begin{table}[t]
\centering
\small
\begin{tabular}{lccc}
\toprule
Model & Text & Layout & Aes. \\
\midrule
OCR expert & \textbf{8.49} & 7.52 & 5.22 \\
Layout expert & 8.42 & \textbf{7.71} & 5.26 \\
Aesthetic expert & 8.35 & 7.33 & \textbf{5.43} \\
Mix-GRPO & 8.38 & 7.51 & 5.30 \\
\textbf{Flow-OPD} & 8.45 & 7.67 & 5.36 \\
\bottomrule
\end{tabular}
\caption{Multi-reward alignment comparison.}
\label{tab:opd}

\end{table}

\textbf{GRPO vs. OPD.}
Table~\ref{tab:opd} isolates multi-reward reconciliation. Single-reward experts reach high scores on their own axes but degrade others. A weighted mixture improves the average but remains below specialist ceilings. OPD approaches all three ceilings simultaneously.

\begin{table}[t]
\centering
\small
\setlength{\tabcolsep}{1.5pt}
\begin{tabular}{lccccc}
\toprule
Dim. & Qwen-T2I & Qwen-CN & Qwen-Edit & \textbf{\method} & Friedman $p$ \\
\midrule
Math & 3.540 & 1.744 & 3.045 & \textbf{1.671} & 4.62e-26 \\
TextAcc & 2.849 & 2.122 & 3.013 & \textbf{2.017} & 5.55e-10 \\
Layout & 2.817 & 2.129 & 3.296 & \textbf{1.758} & 5.46e-11 \\
Aes. & 2.536 & 2.737 & 2.525 & \textbf{2.201} & 2.74e-04 \\
Read. & 3.204 & \textbf{1.764} & 3.029 & 2.003 & 9.16e-24 \\
\bottomrule
\end{tabular}
\caption{Human preference rankings. Lower is better.}
\label{tab:human_rank}
\vspace{-5pt}
\end{table}

\textbf{Human preference.}
Ten participants ranked anonymized outputs from four methods on 80 prompts, including 40 English and 40 Chinese prompts with 20 prompts per chart type. Aesthetic considers visual quality and prompt alignment. As shown in Table~\ref{tab:human_rank} \method{} ranks best on Math, TextAcc, Layout, and Aesthetic. Qwen+CN-open leads Readability mainly because it stays close to plain conventional charts with limited artistic stylization. Friedman tests are significant across all dimensions, and the human rankings generally support the main automatic trends.

\section{Discussion and Limitations}
Despite the strong performance of \method{}, several limitations remain. \method{} still relies on explicit grayscale conditions; it does not prove that pure T2I models can infer chart geometry directly from text. \bench{} focuses on one-dimensional categorical data and four chart types; grouped, stacked, multi-series, scatter, radar, and dashboard settings remain future work. OCR and VLM judges may introduce evaluator bias. Developing more robust, chart-specific evaluation protocols remains an open challenge for the community.

\section{Conclusion}

We present ArtChart, a unified framework encompassing a task definition, benchmark, evaluator, and generation system for artistic chart generation with integrated text rendering. By combining a chart-specific mathematical control module conditioned on text-free grayscale layouts, a multi-reward reinforcement learning strategy optimizing text accuracy, layout quality, and visual aesthetics, and a multi-expert distillation mechanism that resolves inter-reward conflicts, ArtChart jointly addresses the coupled requirements of numerical faithfulness, text correctness, label binding, visual appeal, and readability. ArtChart-Bench and ArtChart-Eval provide a unified testbed for comparing T2I, image-editing, and controllable generation methods, tracking progress toward image generators that produce reliable visual communication rather than merely attractive images.

\clearpage
\appendix

\section{Supplementary materials}
\section{A. Benchmark and Release Notes}

\bench{} contains 2000 evaluation records, with 1000 English prompts and 1000 Chinese prompts. Each language split is balanced across four chart types: vertical bar, horizontal bar, pie, and area. The package also includes a 200-record pilot subset for fast evaluation. \tabref{tab:benchmark-summary} summarizes the submitted benchmark package. The full evaluation code is submitted together with the benchmark records.

Each record stores a natural-language prompt and structured chart data. The natural-language prompt is used by prompt-only systems. Methods that support spatial control additionally receive the deterministic text-free grayscale condition generated by the submitted code. The structured fields include the chart type, title, ordered categories, raw numeric values, display values, style family, and diagnostic tags. These fields are used only by the evaluator.

The benchmark is designed to be useful beyond \method{}. We will publicly release the full evaluation data, evaluation code, and a strong open-source baseline model. The released baseline is not intended to be the final upper bound for this task: closed-source image-generation systems and future pure text-to-image models may continue to improve. Instead, \bench{} provides a stable testbed for measuring whether such systems can preserve mathematical chart structure, rendered text, and label-mark binding under artistic stylization.

\section{B. Training and Implementation Notes}

We use approximately 13K filtered $(P,G,I^\star)$ triples generated from GPT-Image-2 and Nano-Banana-Pro candidates, with an acceptance rate of about 30\% after automatic and human filtering. Training is conducted on $16\times$ A100 GPUs. The ControlNet learning rate is $4{\times}10^{-5}$ and the RL learning rate is $1{\times}10^{-4}$. Quantitative experiments use seed 42 and 40 denoising steps; with Lightning LoRA, the same end-to-end model supports 4-step inference in about 4 seconds per image.

\section{C. OPD Routing Detail}

The main paper describes multi-expert distillation from three single-reward experts: OCR, layout, and aesthetic. The hard router used in OPD is constructed from model behavior on the training prompts. Before OPD distillation, we generate samples on the training prompt-condition pairs and score them with the same three validation signals used for expert selection: text accuracy, text layout, and aesthetics. To avoid scale bias across heterogeneous rewards, each score is converted into a normalized deficit using validation-set statistics. The hard router assigns each sample to the expert corresponding to its largest normalized deficit. Thus, samples with poor text rendering are routed to the OCR expert, samples with label-mark mismatch are routed to the layout expert, and samples with weak visual quality are routed to the aesthetic expert.

This routing strategy avoids manually tuning a static reward mixture for all samples. It also reflects the empirical conflict observed during training: optimizing OCR too aggressively tends to produce rigid chart-like images, while optimizing aesthetics alone can reduce text faithfulness or weaken label-mark binding.

\section{D. Human Preference Study}

Ten participants ranked anonymized outputs from four methods on 80 prompts, including 40 English and 40 Chinese prompts with 20 prompts per chart type. The methods are Qwen-Image-2512, Qwen+CN-open, Qwen-Edit-2511, and \method{}. Annotators ranked the four outputs along five dimensions: mathematical faithfulness, text accuracy, text layout and binding, aesthetic quality, and readability. Ties were allowed. We convert tied rankings into positional ranks before aggregation; for example, a tie for second place gives both tied methods rank 2.5. We first average ranks across annotators for each prompt and then use prompts as the statistical units. \tabref{tab:human-posthoc} reports the post-hoc Wilcoxon signed-rank tests.

The post-hoc results support the automatic trends in the main paper. \method{} is significantly preferred over Qwen-Image-2512 and Qwen-Edit-2511 on all evaluated dimensions and on the overall ranking. Compared with Qwen+CN-open, \method{} is significantly preferred on mathematical faithfulness, text layout and binding, aesthetic quality, and overall preference. Qwen+CN-open is preferred on readability, mainly because it tends to stay closer to a plain conventional chart and therefore sacrifices less readability during artistic transformation.

\begin{figure*}[!t]
\centering
\includegraphics[width=0.98\textwidth]{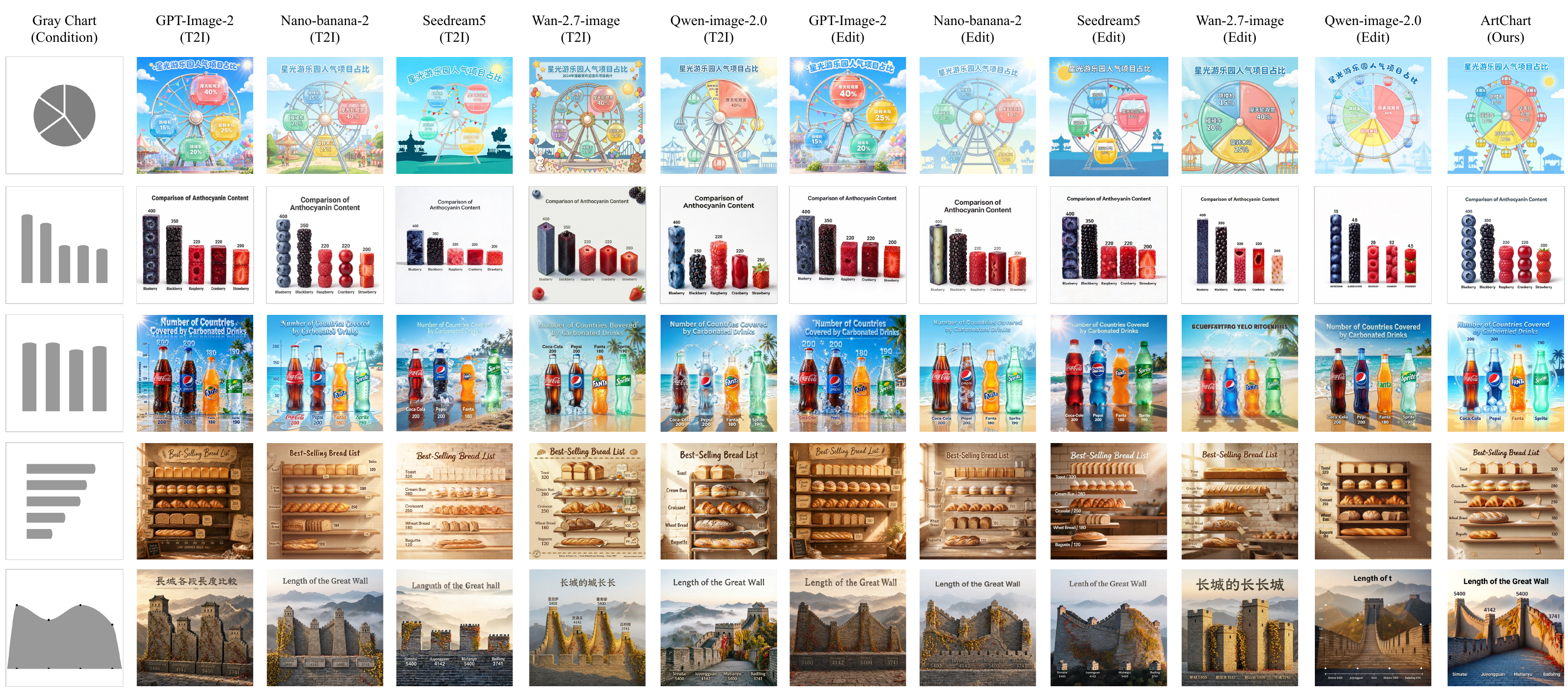}
\caption{Qualitative comparison across multiple methods. Each row starts with the text-free grayscale condition, followed by generated artistic charts from prompt-only or editing methods.}
\label{fig:api-compare}
\end{figure*}

\begin{table}[!t]
\centering
\scriptsize
\resizebox{\columnwidth}{!}{
\begin{tabular}{lcccccc}
\toprule
Method & Math & Text & Layout & Aes. & IF & Read. \\
\midrule
GPT-Image-2(T2I) & \textbf{8.35} & 6.55 & \textbf{9.68} & 5.37 & 9.63 & 8.45 \\
GPT-Image-2(Edit) & 8.28 & 6.99 & 9.49 & 5.26 & \textbf{9.67} & \textbf{8.50} \\
Gemini-3-pro-Image(T2I) & 6.31 & 6.95 & 9.37 & 5.27 & 9.55 & 8.15 \\
Gemini-3-pro-Image(Edit) & 6.70 & 7.68 & 9.15 & 5.28 & 9.55 & 8.03 \\
Qwen-Image-2(T2I) & 4.62 & 8.33 & 7.70 & 5.23 & 8.59 & 6.43 \\
Qwen-Image-2(Edit) & 5.14 & 8.20 & 7.57 & 5.13 & 8.47 & 6.51 \\
Wan2.7-Image-pro(T2I) & 6.12 & 5.49 & 7.48 & 5.28 & 8.85 & 7.42 \\
Wan2.7-Image-pro(Edit) & 7.40 & 7.64 & 7.15 & 5.14 & 8.79 & 7.48 \\
Seedream-5-pro(T2I) & 6.17 & \textbf{8.62} & 8.93 & \textbf{5.47} & 9.43 & 8.06 \\
\bottomrule
\end{tabular}
}
\caption{Quantitative evaluation of API models on the 200-sample \bench{} subset, including both English and Chinese prompts.}
\label{tab:api-subset-all}
\end{table}

\begin{table}[!t]
\centering
\scriptsize
\resizebox{\columnwidth}{!}{
\begin{tabular}{lcccccc}
\toprule
Method & Math & Text & Layout & Aes. & IF & Read. \\
\midrule
GPT-Image-2(T2I) & \textbf{8.66} & 6.50 & \textbf{9.76} & 5.45 & 9.61 & 8.41 \\
GPT-Image-2(Edit) & 8.43 & 7.00 & 9.60 & 5.30 & \textbf{9.66} & \textbf{8.46} \\
Gemini-3-pro-Image(T2I) & 6.45 & 6.79 & 9.36 & 5.33 & 9.52 & 8.16 \\
Gemini-3-pro-Image(Edit) & 7.23 & 7.66 & 9.32 & 5.35 & 9.60 & 8.03 \\
Qwen-Image-2(T2I) & 4.83 & 7.86 & 7.59 & 5.30 & 8.59 & 6.33 \\
Qwen-Image-2(Edit) & 5.76 & 7.79 & 7.52 & 5.27 & 8.49 & 6.56 \\
Wan2.7-Image-pro(T2I) & 6.14 & 4.95 & 7.10 & 5.46 & 8.71 & 7.26 \\
Wan2.7-Image-pro(Edit) & 7.94 & 7.02 & 7.34 & 5.26 & 8.85 & 7.41 \\
Seedream-5-pro(T2I) & 6.97 & \textbf{8.25} & 8.76 & \textbf{5.52} & 9.48 & 7.98 \\
\bottomrule
\end{tabular}
}
\caption{Quantitative results on the Chinese split of the 200-sample \bench{} subset.}
\label{tab:api-subset-zh}
\end{table}

\begin{table}[!t]
\centering
\scriptsize
\resizebox{\columnwidth}{!}{
\begin{tabular}{lcccccc}
\toprule
Method & Math & Text & Layout & Aes. & IF & Read. \\
\midrule
GPT-Image-2(T2I) & 8.08 & 6.60 & \textbf{9.62} & 5.29 & 9.65 & 8.49 \\
GPT-Image-2(Edit) & \textbf{8.11} & 6.99 & 9.37 & 5.22 & \textbf{9.69} & \textbf{8.56} \\
Gemini-3-pro-Image(T2I) & 6.19 & 7.10 & 9.38 & 5.21 & 9.57 & 8.13 \\
Gemini-3-pro-Image(Edit) & 6.21 & 7.69 & 8.98 & 5.21 & 9.51 & 8.02 \\
Qwen-Image-2(T2I) & 4.41 & 8.80 & 7.82 & 5.16 & 8.59 & 6.53 \\
Qwen-Image-2(Edit) & 4.64 & 8.62 & 7.63 & 4.99 & 8.46 & 6.47 \\
Wan2.7-Image-pro(T2I) & 6.10 & 6.04 & 7.84 & 5.10 & 8.99 & 7.57 \\
Wan2.7-Image-pro(Edit) & 6.92 & 8.26 & 6.98 & 5.03 & 8.74 & 7.56 \\
Seedream-5-pro(T2I) & 5.06 & \textbf{9.14} & 9.15 & \textbf{5.39} & 9.36 & 8.16 \\
\bottomrule
\end{tabular}
}
\caption{Quantitative results on the English split of the 200-sample \bench{} subset.}
\label{tab:api-subset-en}
\end{table}

\section{E. Additional Method Comparison}

We also evaluate a broader set of prompt-only and editing systems on a small balanced 200 subset used for closed-source method stress testing. Closed-source API systems are black-box services and may use internal prompting, editing, or agentic pipelines that are not reproducible by users. We therefore treat these results as reference comparisons rather than as direct ablations of our open method. The closed-source API baselines include GPT-Image-2, Gemini-3-pro-Image, Qwen-Image-2, Wan2.7-Image-pro, and Seedream-5-pro. Except for Seedream, we test both text-to-image and editing variants when available. Aggregate results are shown in \tabref{tab:api-subset-all}, \tabref{tab:api-subset-zh}, and \tabref{tab:api-subset-en}. All VLM-based judgments in this subset use Qwen3.5-397B-A17B with thinking mode enabled; disabling thinking mode substantially degrades the reliability of the evaluation.

GPT-Image-2 and Gemini-3-pro-Image obtain particularly high Layout scores, but their Text scores are relatively lower. Visual inspection suggests that a major reason is over-generation of text: these systems often add extra words or decorative annotations beyond the expected chart title, categories, and values. Such extra strings reduce the Text score even when the generated labels are placed in plausible positions. This behavior also supports our treatment of closed-source systems as black-box references, since the final image may be produced from internally rewritten or augmented prompts rather than exactly from the user-provided prompt alone.

\figref{fig:api-compare} highlights two recurring patterns in closed-source API outputs. First, editing APIs do not always faithfully follow the grayscale condition; for example, the pie-chart examples can preserve the topic and style while drifting away from the supplied sector geometry. Second, semantic priors can override data values. In the bread-chart row, several outputs render the baguette as visually long because a baguette is semantically elongated, even though its associated numeric value is short. These cases motivate evaluating artistic charts as visual communication artifacts rather than only as attractive images.

\begin{table}[t]
\centering
\small
\setlength{\tabcolsep}{5pt}
\begin{tabular}{lr}
\toprule
Item & Count or value \\
\midrule
Full evaluation records & 2000 \\
English records & 1000 \\
Chinese records & 1000 \\
Chart types & 4 \\
Records per chart type & 500 \\
Pilot subset records & 200 \\
Style families & 15 \\
Category-count range & 2--10 \\
\bottomrule
\end{tabular}
\caption{High-level summary of the submitted benchmark package.}
\label{tab:benchmark-summary}
\end{table}

\begin{table}[t]
\centering
\scriptsize
\resizebox{\columnwidth}{!}{
\begin{tabular}{lccc}
\toprule
Dim. & Qwen-T2I & Qwen+CN & Qwen-Edit \\
\midrule
Math & $2.070 / 4.96{\times}10^{-11}$ & $0.394 / 3.65{\times}10^{-4}$ & $1.552 / 7.40{\times}10^{-11}$ \\
Text & $0.783 / 4.59{\times}10^{-6}$ & $0.500 / 0.00849$ & $0.946 / 2.07{\times}10^{-7}$ \\
Layout & $1.058 / 1.15{\times}10^{-5}$ & $0.371 / 0.0396$ & $1.538 / 2.12{\times}10^{-8}$ \\
Aes. & $0.335 / 0.00208$ & $0.536 / 9.00{\times}10^{-6}$ & $0.324 / 0.00208$ \\
Read. & $1.201 / 4.09{\times}10^{-10}$ & $-0.239 / 0.0217$ & $1.026 / 1.78{\times}10^{-9}$ \\
Overall & $1.089 / 5.83{\times}10^{-11}$ & $0.213 / 0.00189$ & $1.077 / 5.83{\times}10^{-11}$ \\
\bottomrule
\end{tabular}
}
\caption{Post-hoc Wilcoxon signed-rank tests comparing \method{} with each baseline. Each cell reports mean-rank difference / Holm-corrected $p$-value. Positive differences mean \method{} is preferred; lower rank is better.}
\label{tab:human-posthoc}
\end{table}

\section{G. Closed-Source Results by Chart Type}

We further break down the closed-source subset results by chart type in \tabref{tab:api-subset-chart-type}. Averaged over the nine methods and excluding the overall rows, bar and horizontal-bar charts obtain the strongest Math scores (7.71 and 7.24), while area and pie charts are lower (5.13 and 5.72), reflecting the difficulty of continuous filled regions and sector-angle preservation. Layout remains high for area, bar, and horizontal-bar charts (9.04, 9.00, and 8.95), but drops for pie charts (6.92), where sector-label binding is more fragile. Text accuracy shows a different pattern: pie charts have the highest mean Text score (8.00), followed by horizontal-bar and area charts (7.35 and 7.31), while bar charts are lowest (6.88), suggesting that glyph rendering and semantic label placement fail in different ways.

\begin{table}[!t]
\centering
\scriptsize
\setlength{\tabcolsep}{2pt}
\resizebox{\columnwidth}{!}{
\begin{tabular}{llcccccc}
\toprule
Method & Type & Math & Text & Layout & Aes. & IF & Read. \\
\midrule
\multirow{5}{*}{Gemini-3-pro-Image(Edit)} & Overall & 6.70 & 7.68 & 9.15 & 5.28 & 9.55 & 8.03 \\
 & Area & 4.66 & 8.03 & 9.77 & 5.43 & 9.44 & 7.22 \\
 & Bar & 8.52 & 7.23 & 9.39 & 5.29 & 9.71 & 8.39 \\
 & H-Bar & 7.74 & 7.29 & 9.08 & 5.07 & 9.54 & 7.93 \\
 & Pie & 5.92 & 8.13 & 8.35 & 5.31 & 9.52 & 8.56 \\
\midrule
\multirow{5}{*}{Gemini-3-pro-Image(T2I)} & Overall & 6.31 & 6.95 & 9.37 & 5.27 & 9.55 & 8.15 \\
 & Area & 4.25 & 6.78 & 9.71 & 5.48 & 9.53 & 7.45 \\
 & Bar & 7.91 & 6.81 & 9.77 & 5.17 & 9.55 & 8.52 \\
 & H-Bar & 7.72 & 7.54 & 9.19 & 5.10 & 9.64 & 8.35 \\
 & Pie & 4.05 & 6.70 & 8.39 & 5.31 & 9.41 & 8.25 \\
\midrule
\multirow{5}{*}{GPT-Image-2(Edit)} & Overall & 8.28 & 6.99 & 9.49 & 5.26 & 9.67 & 8.50 \\
 & Area & 7.09 & 7.06 & 9.83 & 5.40 & 9.60 & 8.13 \\
 & Bar & 8.68 & 6.39 & 9.73 & 5.21 & 9.79 & 8.79 \\
 & H-Bar & 8.77 & 6.26 & 9.66 & 5.06 & 9.57 & 7.86 \\
 & Pie & 8.23 & 8.25 & 8.90 & 5.37 & 9.67 & 8.88 \\
\midrule
\multirow{5}{*}{GPT-Image-2(T2I)} & Overall & 8.35 & 6.55 & 9.68 & 5.37 & 9.63 & 8.45 \\
 & Area & 7.91 & 6.29 & 10.00 & 5.56 & 9.82 & 8.04 \\
 & Bar & 8.79 & 5.68 & 9.76 & 5.31 & 9.63 & 8.51 \\
 & H-Bar & 8.21 & 6.21 & 9.78 & 5.20 & 9.65 & 8.27 \\
 & Pie & 8.29 & 8.03 & 9.30 & 5.41 & 9.50 & 8.83 \\
\midrule
\multirow{5}{*}{Qwen-Image-2(Edit)} & Overall & 5.14 & 8.20 & 7.57 & 5.13 & 8.47 & 6.51 \\
 & Area & 3.89 & 8.33 & 8.69 & 5.17 & 8.73 & 5.59 \\
 & Bar & 6.75 & 7.44 & 7.98 & 5.17 & 8.44 & 6.80 \\
 & H-Bar & 6.00 & 8.45 & 8.20 & 4.95 & 8.65 & 6.59 \\
 & Pie & 3.43 & 8.59 & 5.43 & 5.25 & 8.12 & 6.92 \\
\midrule
\multirow{5}{*}{Qwen-Image-2(T2I)} & Overall & 4.62 & 8.33 & 7.70 & 5.23 & 8.59 & 6.43 \\
 & Area & 3.98 & 8.03 & 8.55 & 5.40 & 8.71 & 5.83 \\
 & Bar & 5.18 & 8.19 & 8.04 & 5.18 & 8.53 & 6.79 \\
 & H-Bar & 5.58 & 8.30 & 8.55 & 5.03 & 8.48 & 6.44 \\
 & Pie & 3.37 & 8.80 & 5.05 & 5.30 & 8.69 & 6.60 \\
\midrule
\multirow{5}{*}{Wan2.7-Image-pro(Edit)} & Overall & 7.40 & 7.64 & 7.15 & 5.14 & 8.79 & 7.48 \\
 & Area & 5.65 & 7.64 & 7.96 & 5.24 & 8.58 & 6.25 \\
 & Bar & 8.63 & 7.36 & 7.84 & 5.11 & 8.98 & 8.04 \\
 & H-Bar & 7.37 & 7.65 & 7.83 & 4.92 & 9.02 & 7.86 \\
 & Pie & 6.88 & 7.92 & 4.76 & 5.29 & 8.43 & 7.25 \\
\midrule
\multirow{5}{*}{Wan2.7-Image-pro(T2I)} & Overall & 6.12 & 5.49 & 7.48 & 5.28 & 8.85 & 7.42 \\
 & Area & 4.90 & 5.15 & 7.55 & 5.42 & 8.80 & 6.96 \\
 & Bar & 6.85 & 4.70 & 8.76 & 5.23 & 8.93 & 7.38 \\
 & H-Bar & 6.84 & 5.58 & 8.76 & 5.06 & 8.93 & 7.58 \\
 & Pie & 5.51 & 6.54 & 5.09 & 5.41 & 8.73 & 7.57 \\
\midrule
\multirow{5}{*}{Seedream-5-pro(T2I)} & Overall & 6.17 & 8.62 & 8.93 & 5.47 & 9.43 & 8.06 \\
 & Area & 3.82 & 8.45 & 9.30 & 5.62 & 9.29 & 7.34 \\
 & Bar & 8.08 & 8.12 & 9.71 & 5.41 & 9.63 & 8.45 \\
 & H-Bar & 6.89 & 8.84 & 9.49 & 5.36 & 9.51 & 8.07 \\
 & Pie & 5.78 & 9.08 & 7.03 & 5.44 & 9.26 & 8.37 \\
\bottomrule
\end{tabular}
}
\caption{Per-chart-type breakdown on the 200-sample bench subset.}
\label{tab:api-subset-chart-type}
\end{table}


\begin{thebibliography}{99}

\bibitem[Hunter(2007)]{hunter2007matplotlib}
Hunter, John D. 2007. {Matplotlib}: A 2D Graphics Environment. Computing in Science \& Engineering, 9(3): 90--95.

\bibitem[Bostock, Ogievetsky, and Heer(2011)]{bostock2011d3}
Bostock, Michael; Ogievetsky, Vadim; and Heer, Jeffrey. 2011. {D3}: Data-Driven Documents. IEEE Transactions on Visualization and Computer Graphics, 17(12): 2301--2309.

\bibitem[Li et al.(2018)]{li2018echarts}
Li, Deqing; Mei, Honghui; Shen, Yi; Su, Shuang; Zhang, Wenli; Wang, Junting; Zu, Ming; and Chen, Wei. 2018. {ECharts}: A Declarative Framework for Rapid Construction of Web-Based Visualization. Visual Informatics, 2(2): 136--146.

\bibitem[Dibia(2023)]{dibia2023lida}
Dibia, Victor. 2023. {LIDA}: A Tool for Automatic Generation of Grammar-Agnostic Visualizations and Infographics using Large Language Models. In Proceedings of the Annual Meeting of the Association for Computational Linguistics (ACL).

\bibitem[Maddigan and Susnjak(2023)]{maddigan2023chat2vis}
Maddigan, Paula; and Susnjak, Teo. 2023. {Chat2VIS}: Generating Data Visualisations via Natural Language using {ChatGPT}, {Codex} and {GPT-3} Large Language Models. IEEE Access, 11: 45181--45193.

\bibitem[Zhang, Rao, and Agrawala(2023)]{zhang2023controlnet}
Zhang, Lvmin; Rao, Anyi; and Agrawala, Maneesh. 2023. Adding Conditional Control to Text-to-Image Diffusion Models. In Proceedings of the IEEE/CVF International Conference on Computer Vision (ICCV).

\bibitem[Li et al.(2024)]{li2024controlnetpp}
Li, Xinsir; Hou, Yunfan; Loy, Chen Change; and Cheng, Ming-Ming. 2024. {ControlNet++}: Improving Conditional Controls with Efficient Consistency Feedback. arXiv:2404.07987.

\bibitem[Black et al.(2024)]{black2024ddpo}
Black, Kevin; Janner, Michael; Du, Yilun; Kostrikov, Ilya; and Levine, Sergey. 2024. Training Diffusion Models with Reinforcement Learning. In International Conference on Learning Representations (ICLR).

\bibitem[Fan et al.(2024)]{fan2024dpok}
Fan, Ying; Watkins, Olivia; Du, Yuqing; Liu, Hao; Ryu, Moonkyung; Boutilier, Craig; Abbeel, Pieter; Ghavamzadeh, Mohammad; Lee, Kimin; and Lee, Kuan-Hui. 2024. {DPOK}: Reinforcement Learning for Fine-tuning Text-to-Image Diffusion Models. In Advances in Neural Information Processing Systems.

\bibitem[Qwen Team(2025)]{qwenimage2025}
Qwen Team. 2025. Qwen-Image Technical Report. arXiv:2508.02324.

\bibitem[ERNIE Team(2026)]{ernieimage2026}
ERNIE Team. 2026. ERNIE-Image Technical Report. arXiv:2605.25347.

\bibitem[Xiao et al.(2026)]{xiao2026chartist}
Xiao, Shishi; Zhou, Tongyu; Laidlaw, David H.; and Chan, Gromit Yeuk-Yin. 2026. {ChArtist}: Generating Pictorial Charts with Unified Spatial and Subject Control. In Proceedings of the IEEE/CVF Conference on Computer Vision and Pattern Recognition (CVPR).

\bibitem[Sun et al.(2026)]{sun2026ssalign}
Sun, Zhida; Zhang, Yulin; Gu, Zheng; Lu, Min; Lee, Bongshin; Cohen-Or, Daniel; and Huang, Hui. 2026. Semantic-Structural Alignment for Generative Pictorial Charts. ACM Transactions on Graphics, 45(4).

\bibitem[Liu et al.(2026)]{liu2026showtable}
Liu, Zhihang; Bao, Xiaoyi; Li, Pandeng; Zhou, Junjie; Liao, Zhaohe; He, Yefei; Jiang, Kaixun; Xie, Chen-Wei; Zheng, Yun; and Xie, Hongtao. 2026. {ShowTable}: Unlocking Creative Table Visualization with Collaborative Reflection and Refinement. arXiv:2512.13303.

\bibitem[Tang et al.(2026)]{tang2026igenbench}
Tang, Yinghao; Liu, Xueding; Zhang, Boyuan; Lan, Tingfeng; Xie, Yupeng; Lao, Jiale; Wang, Yiyao; Li, Haoxuan; Gao, Tingting; Pan, Bo; Weng, Luoxuan; Huang, Xiuqi; Zhu, Minfeng; Feng, Yingchaojie; Luo, Yuyu; and Chen, Wei. 2026. {IGenBench}: Benchmarking the Reliability of Text-to-Infographic Generation. arXiv:2601.04498.

\bibitem[Xiao et al.(2024)]{xiao2024chartspark}
Xiao, Shishi; Huang, Suizi; Lin, Yue; Ye, Yilin; and Zeng, Wei. 2024. Let the Chart Spark: Embedding Semantic Context into Chart with Text-to-Image Generative Model. IEEE Transactions on Visualization and Computer Graphics.

\bibitem[Wu, Chung, and Adar(2023)]{wu2023viz2viz}
Wu, Jiaqi; Chung, John Joon Young; and Adar, Eytan. 2023. viz2viz: Prompt-Driven Stylized Visualization Generation Using a Diffusion Model. arXiv:2304.01919.

\bibitem[Yan et al.(2025)]{yan2025charteditor}
Yan, Siyu; Liu, Tiancheng; Yang, Weikai; Tang, Nan; and Luo, Yuyu. 2025. {ChartEditor}: A Human-AI Paired Tool for Authoring Pictorial Charts. arXiv:2501.07320.

\bibitem[Liu et al.(2024)]{liu2024glyphbyt5}
Liu, Rosanne; Garrette, Dan; Saharia, Chitwan; Chan, William; Roberts, Adam; and Lee, Katherine. 2024. {Glyph-ByT5}: A Customized Text Encoder for Accurate Visual Text Rendering. arXiv:2403.09622.

\bibitem[Tuo et al.(2024)]{tuo2024anytext}
Tuo, Yuxiang; Xiang, Wenchang; He, Jun-Yan; Geng, Yifeng; and Xie, Xuansong. 2024. {AnyText}: Multilingual Visual Text Generation and Editing. arXiv:2404.14249.

\bibitem[Chen et al.(2023)]{chen2023textdiffuser2}
Chen, Jingye; Huang, Yupan; Lv, Tengchao; Cui, Lei; Chen, Qifeng; and Wei, Furu. 2023. {TextDiffuser-2}: Unleashing the Power of Language Models for Text Rendering. arXiv:2311.16465.

\bibitem[Liu et al.(2025a)]{liu2025easytext}
Liu, Xueding; Tang, Yinghao; Huang, Xiuqi; Luo, Yuyu; and Chen, Wei. 2025. {EasyText}: Controllable Diffusion Transformer for Multilingual Text Rendering. arXiv:2505.24417.

\bibitem[Zhu et al.(2026)]{zhu2026textpecker}
Zhu, Hanshen; Liu, Yuliang; Wu, Xuecheng; Wang, An-Lan; Feng, Hao; Yang, Dingkang; Feng, Chao; Huang, Can; Tang, Jingqun; and Bai, Xiang. 2026. {TextPecker}: Rewarding Structural Anomaly Quantification for Enhancing Visual Text Rendering. arXiv:2602.20903.

\bibitem[Cui et al.(2026)]{cui2026textalign}
Cui, Mingxuan; Yang, Jingpu; Ji, Fengxian; Jiang, Qian; Shi, Zhecheng; Wang, Jiaming; Song, Zirui; Koto, Fajri; and Chen, Xiuying. 2026. {TextAlign}: Preference Alignment for Text Rendering with Hierarchical Rewards. arXiv:2605.19320.

\bibitem[Liu et al.(2025b)]{liu2025flowgrpo}
Liu, Jie; Liu, Gongye; Liang, Jiajun; Li, Yangguang; and Liu, Jiaheng. 2025. {Flow-GRPO}: Training Flow Matching Models via Online Reinforcement Learning. arXiv:2505.05470.

\bibitem[Fan et al.(2026)]{fan2026poca}
Fan, Yaohou; Wang, Qingzhong; Huang, Yongsong; Liu, Junyi; Miyazaki, Tomo; and Omachi, Shinichiro. 2026. {POCA}: Pareto-Optimal Curriculum Alignment for Visual Text Generation. arXiv:2604.24171.

\bibitem[Fang et al.(2026)]{fang2026flowopd}
Fang, Zhen; Huang, Wenxuan; Zeng, Yu; Zhao, Yiming; Chen, Shuang; Feng, Kaituo; Lin, Yunlong; Chen, Lin; Chen, Zehui; Cao, Shaosheng; and Zhao, Feng. 2026. {Flow-OPD}: On-Policy Distillation for Flow Matching Models. arXiv:2605.08063.

\bibitem[Kirstain et al.(2023)]{kirstain2023pickscore}
Kirstain, Yuval; Polyak, Adam; Singer, Uriel; Matiana, Shahbuland; Penna, Joe; and Levy, Omer. 2023. {Pick-a-Pic}: An Open Dataset of User Preferences for Text-to-Image Generation. In Advances in Neural Information Processing Systems.

\bibitem[Masry et al.(2022)]{masry2022chartqa}
Masry, Ahmed; Long, Do Xuan; Tan, Jia Qing; Joty, Shafiq; and Hoque, Enamul. 2022. {ChartQA}: A Benchmark for Question Answering about Charts with Visual and Logical Reasoning. In Findings of the Association for Computational Linguistics: ACL.

\bibitem[Yang et al.(2024)]{yang2024matplotbench}
Yang, Zhou; Li, Yifan; Wang, Jiaqi; and Chen, Wei. 2024. {MatplotBench}: Evaluating Multimodal Models on Scientific Plot Generation. arXiv:2407.00981.

\bibitem[Discus0434(2024)]{aestheticpredictor}
Discus0434. 2024. {Aesthetic Predictor}: SigLIP-Based Aesthetic Score Predictor. \url{https://github.com/discus0434/aesthetic-predictor-v2-5}.

\end{thebibliography}
\end{document}